\documentclass[runningheads]{llncs}
\usepackage{esvect}
\usepackage{float}
\usepackage{amsmath}
\usepackage{amssymb}
\usepackage{booktabs}
\usepackage{graphicx}
\usepackage{hyperref}
\usepackage{url}
\usepackage[numbers]{natbib}
\usepackage{esvect}
\usepackage{amsmath}
\usepackage{amssymb}
\usepackage{mathtools}
\usepackage{bm}
\usepackage{microtype}
\usepackage{multirow}
\usepackage{float}
\usepackage[table]{xcolor}
\usepackage{subcaption}
\usepackage[numbers]{natbib}
\usepackage{wrapfig}
\usepackage{pgfplots}
\usepgfplotslibrary{groupplots}
\pgfplotsset{compat=1.18}
\usepackage[compatibility=false]{caption}
\usepackage{xr}
\usepackage{hyperref}

\title{Unsupervised Anomaly Detection Using Flow Matching on Tabular Data}
\author{Philip Konz\inst{1} \and Tejaswini Medi\inst{1} \and Margret Keuper\inst{1,2}}
\authorrunning{Konz et al.}
\institute{University of Mannheim$^{1}$, MPI for Informatics, Saarland Informatics Campus$^{2}$, Germany\\
\email{philip.konz@gmx.de}}

\begin{document}

\maketitle

\begin{abstract}
Financial anomaly detection often relies on large unlabeled transaction logs,
where anomalous samples may already be present during training. This
training-set contamination violates the clean-normal data assumption used by many
anomaly detection methods. Although flow matching has shown strong performance
in generative modeling, its robustness for unsupervised tabular anomaly detection
remains underexplored. In this work, we study flow-matching-based anomaly
detection under contaminated training set by comparing Time-Conditioned Contraction
Matching (TCCM) with Forest-Flow and evaluating multiple anomaly scoring
functions. Our results show that the anomaly scoring function is critical: TCCM's original single-step Decision score is sensitive to contamination, whereas trajectory-based Deviation and Reconstruction scores provide more stable anomaly signals, such that Forest-Flow becomes competitive with, and in some cases better than TCCM. Our findings highlight the importance of anomaly scoring for
flow-matching in financial anomaly detection under severe anomaly class imbalance.

\end{abstract}

\section{Introduction}

Anomaly detection on tabular data is critical in high-stakes domains such as financial fraud detection \cite{Hilal2022}, manufacturing fault detection \cite{YuZhang2023}, and medical diagnosis \cite{Fernando2021}, where rare abnormal samples often correspond to costly or safety-critical events. Existing methods are broadly categorized into classical machine learning and deep learning approaches \cite{li2025scalable}. Classical methods, including Local Outlier Factor (LOF) \cite{Breunig2000LOF}, Principal Component Analysis (PCA) \cite{Shyu2003}, and one-class Support Vector Machines (SVMs) \cite{Schoelkopf1999}, are widely used because of their simplicity but often struggle with high-dimensional and large-scale datasets due to limited scalability and reduced effectiveness in complex feature spaces \cite{li2025scalable}. Deep learning methods overcome some of these limitations by learning richer representations and have demonstrated strong empirical performance \cite{pangnew}. However, adversarial models such as AnoGAN \cite{schlegl2017unsupervised} suffer from unstable min--max optimization, diffusion-based methods \cite{livernoche2023diffusion} require carefully designed noise schedules and slow iterative inference, and normalizing flows are constrained by invertibility requirements and Jacobian computations \cite{maziarka2021oneflow}. Many deep anomaly detectors also incur high inference latency, limiting their practicality on large-scale tabular data \cite{li2025scalable}.

Another important limitation is that both classical and deep learning methods are typically evaluated in semi-supervised settings assuming clean normal training data \cite{li2025scalable}. In practice, however, training sets often contain unlabeled anomalies, and identifying or removing them is expensive \cite{huang2024entropystop}. Such contamination can substantially degrade detection performance \cite{huang2024entropystop}. Recent work has therefore explored robust strategies such as ensemble-based detectors, which reduce the influence of contaminated samples by aggregating diverse models \cite{huang2024entropystop}, albeit at increased computational cost.

Flow Matching \cite{lipman2022flow} has recently emerged as a powerful generative modeling framework that combines the stability and expressivity of diffusion models with more efficient deterministic sampling dynamics \cite{liu2022flow_straight_fast,lee2023minimizing_trajectory_curvature,li2025scalable}. It learns a time-dependent vector field that continuously transports samples from a simple source distribution, such as Gaussian noise, to the target data distribution. Building on this framework, Time-Conditioned Contraction Matching (TCCM) adapts flow matching for tabular anomaly detection and achieves state-of-the-art performance in semi-supervised settings \cite{li2025scalable}. Instead of learning the full transport vector field, TCCM predicts a time-conditioned contraction vector that progressively moves samples toward the origin.

Despite its strong performance, TCCM is designed for clean normal training data, leaving its behavior under fully unsupervised settings with contaminated training sets largely unexplored. Moreover, although the original work reports weaker performance for conventional flow-matching models, this comparison is limited to the semi-supervised setting. Consequently, the robustness of flow-matching-based anomaly detectors under training-set contamination remains insufficiently understood.

In this work, we investigate flow-matching-based anomaly detection in the fully unsupervised setting. We first examine how different anomaly scoring functions affect the robustness of TCCM under contaminated training data and whether alternative scoring strategies mitigate performance degradation. We then evaluate whether conventional flow-matching models, specifically ForestFlow~\cite{jolicoeur2024forestflow}, can match or outperform TCCM when paired with suitable scoring functions. Experiments are conducted primarily on two financial tabular datasets with markedly different anomaly ratios: Campaign ($\rho = 11.27\%$) and Synthetic Business Transaction ($\rho = 0.786\%$), enabling evaluation under both moderate and extreme class imbalance.

Our main contributions are as follows:
\begin{itemize}
    \item We investigate flow-matching-based anomaly detection under contaminated training data, focusing on realistic unsupervised financial tabular datasets with unlabeled anomalies.
    \item We show that anomaly scoring is crucial for robustness: TCCM's original single-step Decision score is highly sensitive to contamination, whereas trajectory-based scores, namely Deviation and Reconstruction, provide more stable anomaly signals.
    \item We demonstrate that trajectory-based scoring substantially improves robustness and detection performance. In particular, Deviation and Reconstruction scores enable ForestFlow to become competitive with, and in some cases outperform, the state-of-the-art TCCM baseline across both moderate and severe anomaly class imbalance settings.
\end{itemize}

\paragraph{Our work is organized as follows. Section~2 introduces the proposed models and anomaly scoring functions. Section~3 describes the experimental setup. Section~4 presents the main results, Section~5 analyzes hyperparameter sensitivity, and Section~6 concludes the findings.}

\section{Methodology}

We evaluate two flow-matching-based models with distinct design goals:
Forest-Flow~\citep{jolicoeur2024forestflow}, a general-purpose generative model
for tabular data, and TCCM~\citep{li2025scalable}, a flow-based approach designed
for tabular anomaly detection. Although both learn ordinary differential equation
(ODE)-based velocity fields via flow matching, they differ in how they represent
normality and derive anomaly scores. Forest-Flow learns a generative transport
from Gaussian noise to the data distribution, with anomaly detection performed
post hoc by measuring how well a sample conforms to the learned generative
dynamics. In contrast, TCCM learns a contraction field that pushes samples
toward a fixed point, making deviations from the learned contraction behavior a
direct anomaly signal.

For both models, we evaluate three anomaly scoring functions:
Decision, Deviation, and Reconstruction. The Decision score corresponds to the original TCCM scoring function,
which is Lipschitz-continuous with respect to the
input~\citep{li2025scalable}. It evaluates the velocity prediction error at a
single representative time point: the contraction prediction error at the
original input for TCCM, and the transport velocity prediction error near the
data endpoint for Forest-Flow. The Deviation score accumulates
velocity prediction errors across multiple time steps along the learned flow
trajectory, measuring the temporal consistency of the learned dynamics. The Reconstruction score numerically integrates the learned velocity field
using Euler steps and measures the $\ell_2$ reconstruction error at the
expected endpoint, capturing errors that accumulate along the flow trajectory.

\paragraph{Notation.}
Let $x \in \mathbb{R}^d$ denote a tabular sample, where $d$ is the feature
dimension, and let $z \sim \mathcal{N}(0,I_d)$ denote standard Gaussian noise.
The time variable is $t \in [0,1]$, discretized as
$t_i = i/(n_t-1)$ for $i=0,\ldots,n_t-1$. For Monte Carlo estimation in
Forest-Flow, we draw $K$ independent noise samples $z^{(k)}$,
$k=1,\ldots,K$, for each input sample.

\subsection{Forest-Flow with XGBoost Regressor (FF-XGB)}
Forest-Flow~\citep{jolicoeur2024forestflow} learns a velocity field under the
conditional flow-matching framework~\citep{lipman2022flow}. During training,
each data point $x$ is paired with Gaussian noise $z$, and intermediate states
are constructed by linear interpolation, $x_t = (1-t)z + tx$. The corresponding
target velocity is the constant displacement
$v_{\mathrm{true}} = x-z$, which defines the transport direction from noise to
data. Forest-Flow discretizes time into $n_t$ levels and trains a separate
XGBoost regressor~\citep{chen2016xgboost} $v_\theta^{(i)}$ at each time step
$t_i$ to predict this displacement from $x_{t_i}$. The training objective is
\[
\mathbb{E}\left[
\left\|v_\theta^{(i)}(x_{t_i}) - (x-z)\right\|_2^2
\right],
\]
which encourages the learned velocity field to align with the conditional
transport path. At test time, each sample $x$ is paired with $K$ independent
noise samples $z^{(k)}$, yielding interpolated points
$x_{t_i}^{(k)} = (1-t_i)z^{(k)} + t_i x$. Since later time steps lie closer to
the data distribution and are less dominated by noise, we evaluate anomaly
scores on
\[
\mathcal{T}_{\mathrm{FF}}
=
\left\{
t_i : i=\lfloor n_t/2 \rfloor,\ldots,n_t-2
\right\}.
\]
The Decision score measures the velocity prediction error at the last evaluated
time step. The Deviation score accumulates velocity prediction errors across
all $t_i \in \mathcal{T}_{\mathrm{FF}}$. The Reconstruction score measures how
accurately $x$ can be recovered by numerically integrating the learned velocity
field using Euler steps. Scores are averaged over the $K$ noise realizations to
obtain the final anomaly score for each sample. Formal definitions of the
scoring functions are provided in Appendix~\autoref{app:scoring}.

\subsection{Time-Conditioned Contraction Matching (TCCM) } 

TCCM~\citep{li2025scalable} is not a generative noise-to-data model like
Forest-Flow. Instead, it learns a deterministic contraction field that moves
samples toward the origin. Given a sample $x$ and time $t \in [0,1]$, a single
multi-layer perceptron (MLP) $f_\theta$ predicts a time-conditioned velocity
using sinusoidal time embeddings. The training objective is
\[
\mathbb{E}_{x,t}
\left[
\left\|
f_\theta\big([x;\mathrm{Embed}(t)]\big) + x
\right\|_2^2
\right],
\]
which encourages the predicted velocity to match the contraction direction
$-x$. Thus, the origin acts as a stable fixed point of the learned dynamics.

Unlike Forest-Flow, TCCM uses one continuously time-conditioned network, does
not learn a generative transport from noise to data, and does not require Monte
Carlo sampling. For evaluation, we use the deterministic probe path
$x(t_i) = (1-t_i)x$ and evaluate early time levels
\[
\mathcal{T}_{\mathrm{TCCM}}
=
\left\{
t_i : i = 0,\ldots,\lfloor n_t/2 \rfloor - 1
\right\},
\]
where probe points remain close to the original sample. The Decision score
measures contraction prediction error at the original input. The Deviation
score accumulates prediction errors along the probe path. The Reconstruction
score measures the error after numerically integrating the learned contraction
field toward the origin. Formal definitions of the scoring functions are
provided in Appendix~\autoref{app:scoring}.

\section{Experiments}
\vspace{-0.05cm}
\subsection{Datasets} We evaluate the models on three tabular anomaly detection datasets with
different anomaly ratios and data characteristics. 

\textbf{Campaign} is a public
financial dataset from ADBench~\citep{han2022adbench}, with 41{,}188 samples,
62 numerical features, and 4{,}640 labeled anomalies, corresponding to an
anomaly ratio of $\rho = 11.27\%$. This high-dimensional dataset serves as our
primary financial benchmark and was also used in the original TCCM
evaluation~\citep{li2025scalable}.

\textbf{Synthetic Business Transaction} is a synthetic financial transaction
dataset modeling business-to-business (B2B) payment activity for nine customers.
It contains 6{,}364 transactions with 50 labeled anomalies, corresponding to an
anomaly ratio of $\rho = 50/6364 \approx 0.786\%$. Each transaction is encoded
using 21 numerical, temporally derived features. The anomalies cover suspicious
patterns such as amount spikes, duplicate payments, payee or IBAN changes,
temporal shifts, and atypical payment channels. This makes the dataset suitable
for evaluating anomaly detection under extreme anomaly class imbalance. Further
details are provided in Appendix~\autoref{app:synthetic_dataset}.

To assess generalization beyond financial data, we also evaluate
\textbf{Waveform}, a physics dataset from ADBench~\citep{han2022adbench} that
was used in the original TCCM evaluation~\citep{li2025scalable}. It contains
3{,}443 samples, 21 numerical features, and 100 anomalies, corresponding to anomaly ratio of
$\rho = 2.90\%$.

\subsection{Evaluation Protocol}

We evaluate robustness under training-set contamination following
\citep{li2025scalable}. For \textbf{Campaign}, normal samples are randomly
split into equally sized training and test sets using seed 42. Anomalies are
split evenly, with one half added to the fixed test set and the other half
reserved for controlled injection into the training set. For \textbf{Synthetic Business Transaction}, we use a chronological midpoint
split to preserve its time-series structure. Because anomalies are temporally
non-uniform, this yields 28 anomalies in the training period and 22 in the test
period. Features are z-score normalized using statistics estimated from the
training set only and then applied to the test set to avoid information
leakage~\citep{TCCMCode}.

We evaluate contamination levels of the training set from $0.0\%$ up to each dataset's anomaly ratio. Each model is trained with five random seeds $(0$--$4)$, while
the test set remains fixed. Performance is reported using Area Under the Receiver Operating Characteristic Curve (AUROC) and Area Under the Precision-Recall Curve (AUPRC)
\citep{li2025scalable,mcdermott2024closer}. AUROC measures anomaly detection ranking quality, with $1.0$ being the perfect
discrimination. AUPRC summarizes the precision--recall trade-off and is
especially informative under severe anomaly class imbalance. Higher values indicate
better performance for both metrics.

\subsection{Hyperparameter Details}

\textbf{Forest-Flow.}
Following the large-dataset setting ($N>10{,}000$) in~\citep{jolicoeur2024forestflow}, we use $n_t=20$ time steps and $K=20$ Monte Carlo noise samples per data point for both training and scoring. Each time-specific XGBoost regressor uses the default hyperparameters: $n_{\mathrm{estimators}}=100$, $\mathrm{max\_depth}=7$, $\mathrm{eta}=0.3$, and $\mathrm{tree\_method}=\texttt{hist}$.

\textbf{TCCM.}
Following~\citep{li2025scalable}, we use a three-layer MLP with 256 hidden units per layer, ReLU activations, sinusoidal time embeddings~\citep{vaswani2017attention}, and Adam optimization. We adopt the recommended hyperparameters for each dataset: Campaign uses 50 epochs, batch size 1024, and learning rate 0.005; Waveform uses 580 epochs, batch size 512, and learning rate 0.005. As Synthetic Business Transaction is not evaluated in~\citep{li2025scalable}, we use the default configuration of 100 epochs, batch size 64, and learning rate 0.001. For Deviation and Reconstruction scoring, we set $n_t=20$ to match Forest-Flow.

\section{Evaluations}

\begin{table*}[t]
\centering
\footnotesize
\setlength{\tabcolsep}{4pt}
\renewcommand{\arraystretch}{1.05}
\caption{\small Comparison of TCCM and FF-XGB on three tabular anomaly detection datasets using Decision (Dec.), Deviation (Dev.), and Reconstruction (Rec.) scores. AUROC and AUPRC are reported under zero (${}_0$) and full (${}_{\mathrm{full}}$) training contamination as mean$\pm$std over five runs. Best results are shown in \textbf{bold}.}
\label{tab:four_dataset_compact_results}
\resizebox{\textwidth}{!}{
\begin{tabular}{@{}llcccc@{}}
\toprule
\textbf{Dataset} & \textbf{Method-Score}
& \textbf{AUROC$_0$}
& \textbf{AUROC$_{\mathrm{full}}$}
& \textbf{AUPRC$_0$}
& \textbf{AUPRC$_{\mathrm{full}}$} \\
\midrule

\multirow{6}{*}{Campaign}
& TCCM-Dec.   & $0.778\pm0.025$ & $0.763\pm0.013$ & $0.334\pm0.027$ & $0.309\pm0.012$ \\
& FF-XGB-Dec. & $0.767\pm0.003$ & $0.698\pm0.004$ & $0.297\pm0.004$ & $0.216\pm0.003$ \\
& TCCM-Dev.   & $0.750\pm0.036$ & $0.754\pm0.031$ & $0.323\pm0.031$ & $0.299\pm0.017$ \\
& FF-XGB-Dev. & $0.765\pm0.004$ & $0.729\pm0.002$ & $0.327\pm0.003$ & $0.266\pm0.002$ \\
& TCCM-Rec.   & $0.764\pm0.003$ & $0.764\pm0.002$ & $0.337\pm0.002$ & $\mathbf{0.336\pm0.002}$ \\
& FF-XGB-Rec. & $\mathbf{0.820\pm0.001}$ & $\mathbf{0.804\pm0.002}$ & $\mathbf{0.341\pm0.003}$ & $0.315\pm0.002$ \\

\midrule

\multirow{6}{*}{B2B}
& TCCM-Dec.   & $0.697\pm0.008$ & $0.488\pm0.011$ & $0.037\pm0.008$ & $0.007\pm0.000$ \\
& FF-XGB-Dec. & $0.662\pm0.019$ & $0.523\pm0.011$ & $0.037\pm0.024$ & $0.008\pm0.000$ \\
& TCCM-Dev.   & $0.679\pm0.011$ & $0.476\pm0.009$ & $0.033\pm0.008$ & $0.007\pm0.000$ \\
& FF-XGB-Dev. & $0.777\pm0.003$ & $0.686\pm0.009$ & $\mathbf{0.195\pm0.011}$ & $0.013\pm0.001$ \\
& TCCM-Rec.   & $\mathbf{0.869\pm0.003}$ & $\mathbf{0.846\pm0.001}$ & $0.074\pm0.001$ & $0.064\pm0.001$ \\
& FF-XGB-Rec. & $0.825\pm0.012$ & $0.789\pm0.018$ & $0.059\pm0.003$ & $\mathbf{0.094\pm0.004}$ \\

\midrule

\multirow{6}{*}{Waveform}
& TCCM-Dec.   & $0.677\pm0.067$ & $0.596\pm0.057$ & $0.056\pm0.012$ & $0.042\pm0.007$ \\
& FF-XGB-Dec. & $0.691\pm0.017$ & $0.614\pm0.012$ & $0.069\pm0.007$ & $0.046\pm0.004$ \\
& TCCM-Dev.   & $0.676\pm0.089$ & $0.610\pm0.092$ & $0.068\pm0.021$ & $0.052\pm0.017$ \\
& FF-XGB-Dev. & $0.735\pm0.008$ & $0.678\pm0.012$ & $0.081\pm0.005$ & $0.060\pm0.004$ \\
& TCCM-Rec.   & $0.613\pm0.027$ & $0.616\pm0.009$ & $0.043\pm0.002$ & $0.043\pm0.001$ \\
& FF-XGB-Rec. & $\mathbf{0.741\pm0.017}$ & $\mathbf{0.706\pm0.018}$ & $\mathbf{0.156\pm0.010}$ & $\mathbf{0.102\pm0.014}$ \\

\bottomrule
\end{tabular}
}
\end{table*}

Table~\ref{tab:four_dataset_compact_results} compares TCCM~\citep{li2025scalable}
with Forest-Flow~\citep{jolicoeur2024forestflow} using an
XGBoost~\citep{chen2016xgboost} regressor on Campaign, Synthetic Business
Transaction (B2B), and Waveform. For each method, we evaluate Decision (Dec.),
Deviation (Dev.), and Reconstruction (Rec.) anomaly scores, and report AUROC
and AUPRC under zero-contamination training, where the training set contains
only normal samples, and full-contamination training, where all available
training anomalies of anomaly ratio specific to datasets are included. The test set remains fixed in both settings.

The results in Table~\ref{tab:four_dataset_compact_results} show that anomaly score selection strongly affects robustness under
training-set contamination. On Campaign, FF-XGB-Rec. achieves the best AUROC
under both zero and full contamination ($0.820$ and $0.804$), as well as the
best zero-contamination AUPRC ($0.341$), while TCCM-Rec. gives the best
full-contamination AUPRC ($0.336$). On the highly imbalanced B2B dataset,
TCCM-Rec. obtains the best AUROC under both zero and full contamination
($0.869$ and $0.846$). FF-XGB-Dev. achieves the best zero-contamination AUPRC
($0.195$), whereas FF-XGB-Rec. gives the best full-contamination AUPRC
($0.094$). On Waveform, FF-XGB-Rec. performs best across all metrics, achieving
AUROC values of $0.741$ and $0.706$ and AUPRC values of $0.156$ and $0.102$
under zero and full contamination, respectively.

Overall, trajectory-based scores, particularly Reconstruction, provide more stable anomaly signals than the single-step Decision score. FF-XGB-Rec. performs best on Campaign and Waveform, while TCCM-Rec. achieves the highest AUROC on the B2B dataset. A possible explanation is that the Decision score depends on the prediction error at a single evaluation point, so any local bias introduced by contaminated training data directly affects the anomaly score. By contrast, the Deviation and Reconstruction scores accumulate prediction errors along the learned trajectory. If contamination primarily affects only parts of the learned vector field, aggregating information over multiple time steps reduces the influence of these local deviations and emphasizes the overall consistency of the learned dynamics. This makes trajectory-based scores more robust to training-set contamination. Additional contamination curves across all contamination levels are provided in Appendix~\autoref{contamination}.

\section{Hyperparameter Sensitivity}

\begin{figure}[t]
    \centering
    \includegraphics[
        width=\linewidth,
        trim=5 5 5 5,
        clip
    ]{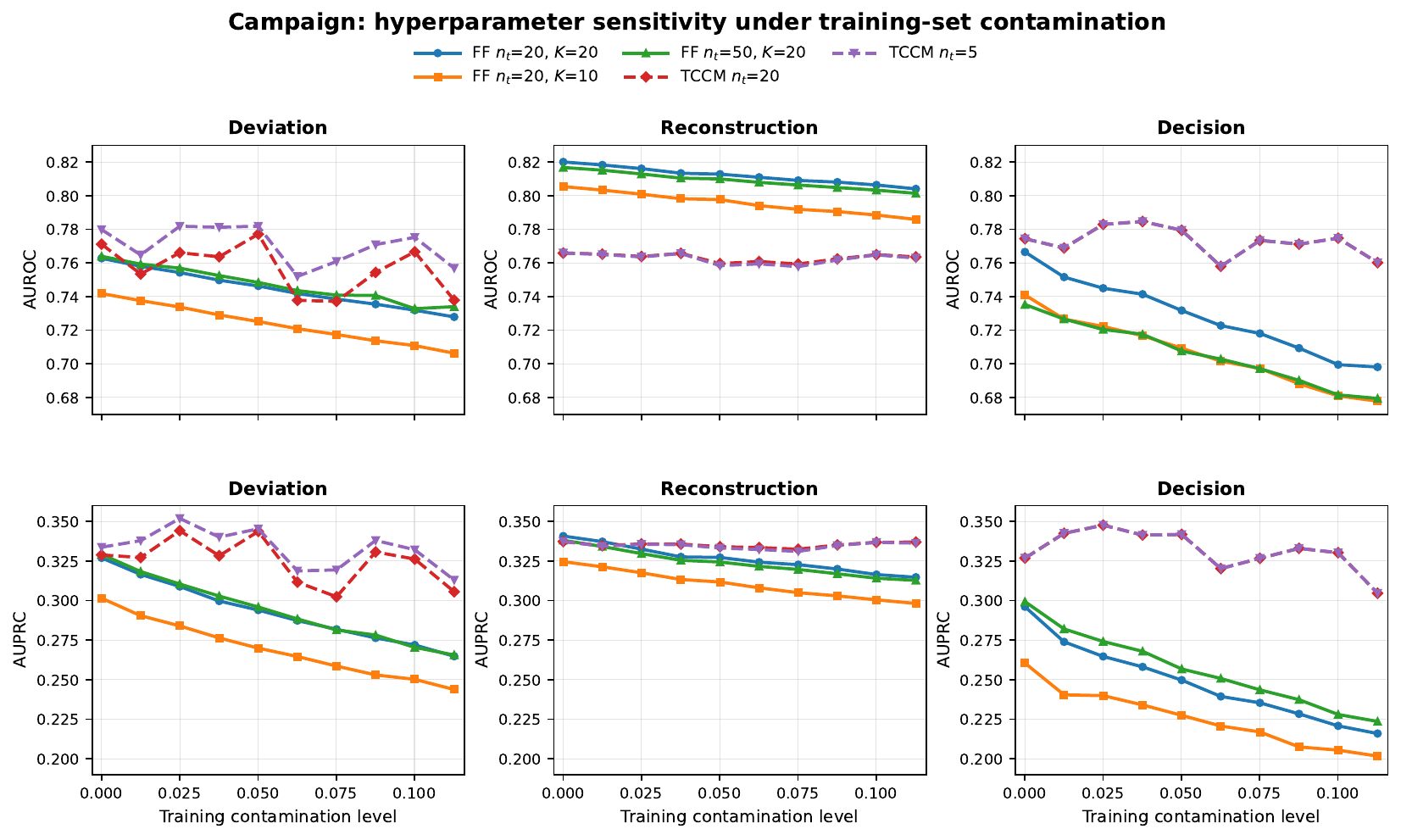}
    \caption{
    Hyperparameter sensitivity on the Campaign dataset under increasing
    training-set contamination. AUROC and AUPRC are reported for the
    Deviation, Reconstruction, and Decision scores. For Forest-Flow,
    \(n_t\) denotes the number of time steps and \(K\) the number of
    Monte Carlo noise samples per data point.
    For TCCM, only \(n_t\) is varied because the model does not use
    Monte Carlo noise-sample averaging.
    }
    \label{fig:hyperparameter_sensitivity}
\end{figure}

Figure~\ref{fig:hyperparameter_sensitivity} summarizes the effect of the hyperparameters on AUROC and AUPRC across contamination levels. For Forest-Flow, increasing the number of noise samples ($K$) generally improves anomaly detection performance, with the largest gains for the Decision score. Reconstruction is less sensitive to $K$, suggesting that trajectory-based aggregation already provides stable anomaly estimates. Compared with the default setting ($K=20$), using $K=10$ (Figure~\ref{fig:hyperparameter_sensitivity}) degrades the performance of the Decision and Deviation scores. In contrast, increasing the number of time steps ($n_t$) from 20 to 50 has only a minor effect, particularly for Deviation and Reconstruction.

For TCCM, reducing $n_t$ from 20 to 5 improves the stability and mean performance of the Deviation score, whereas the Decision score is unaffected because it is always evaluated at $t=1$, irrespective of $n_t$. For Deviation, a larger $n_t$ extends the probe trajectory to increasingly shrunk points relative to the training inputs, introducing prediction noise that weakens the anomaly signal. A smaller $n_t$ keeps the trajectory closer to the training distribution, improving robustness. Reconstruction is less affected because the finer Euler integration associated with a larger $n_t$ partially offsets the impact of the more off-distribution probe points.

Overall, the Reconstruction score is the most stable across hyperparameter settings and contamination levels for both Forest-Flow and TCCM, consistently achieving strong AUROC and AUPRC. In contrast, the Decision score is the most sensitive to hyperparameter choices. 

\section{Conclusion}

We studied flow-matching-based unsupervised anomaly detection under contaminated tabular training data, with a focus on financial applications where unlabeled anomalies may be present in transaction logs. Our results show that robustness depends not only on the underlying flow-matching model but also on the choice of anomaly score. While the original TCCM anomaly score is sensitive to training-set contamination, trajectory-based scores such as Deviation and Reconstruction provide more robust anomaly signals by aggregating velocity prediction errors along the learned flow trajectory. Moreover, Forest-Flow, originally proposed for tabular data generation, becomes a competitive anomaly detector when combined with trajectory-based scoring. These findings highlight the importance of anomaly score design when adapting flow-matching models for anomaly detection. Future work will investigate larger financial benchmarks and more computationally efficient trajectory-based scoring methods.

\clearpage

\appendix
\section*{Appendix} \addcontentsline{toc}{section}{Supplementary Material}

\section{Scoring Function Definitions}
\label{app:scoring}

This section describes the two flow matching frameworks formally. For each model, we describe the underlying framework and define three anomaly scoring functions for each model.

\paragraph{Notation.} Let $x \in \mathbb{R}^d$ denote a data sample and $z \sim \mathcal{N}(0, I_d)$ a standard Gaussian noise vector, where $d$ is the number of feature dimensions.
The time parameter $t$ indexes the transformation process. Discrete time levels are denoted $t_i$ for $i = 0, 1, \ldots, n_t - 1$, where $n_t$ is the number of time levels. To obtain the expected values (Monte-Carlo estimate), we sample $K$ (duplication factor) independent noise vectors $z^{(k)} \sim \mathcal{N}(0, I_d)$ for $k = 1, \ldots, K$ to build interpolations with a test or training sample. $\theta$ denotes the set of learnable model parameters (e.g., XGBoost trees or neural network weights).
The notation is consistent across different scores for one model.

\subsection{Forest-Flow}

\label{subsec:Forest-Flow}
Forest-Flow~\cite{jolicoeur2024forestflow} is a tree-based implementation of conditional flow matching \cite{lipman2022flow} originally proposed for tabular data generation. We adapt it for anomaly detection by training the model and evaluating three scoring functions. Note that the actual implementation of the Forest-Flow code \cite{ForestDiffusionCode} and the paper \cite{jolicoeur2024forestflow} deviate from their time setting and the defined noise levels. We take the framework of code implementation \cite{ForestDiffusionCode} as the scoring functions are adapted to it. Very minor details of the code implementation are disregarded for simplicity (i.e. numerical stability constant for time).

\subsubsection{Framework}
Forest-Flow learns a time-dependent vector field that transports (scaled) samples from a standard Gaussian distribution to the data distribution \cite{ForestDiffusionCode}. It uses a linear interpolation path between noise $z \sim \mathcal{N}(0, I_d)$ (at $t = 0$) and data $x$ (at $t = 1$):
\begin{equation}
    x_t = (1 - t) \cdot z + t \cdot x, \quad t \in [0, 1]
    \label{eq:Forest-Flow_interpolation}
\end{equation}
The true velocity (training target) along this linear path is constant as proposed by \cite{lipman2022flow}:
\begin{equation}
    v_{\text{true}} = x - z
    \label{eq:Forest-Flow_velocity}
\end{equation}

Forest-Flow trains a separate XGBoost regressor for each discrete time level $t_i$ \cite{jolicoeur2024forestflow}, uniformly spaced in $[0, 1]$:
\begin{equation}
    t_i = \frac{i}{n_t - 1}, \quad i = 0, 1, \ldots, n_t - 1
    \label{eq:Forest-Flow_timelevels}
\end{equation}
This results in $n_t$ models, where each model (here the XGBoosts; with $\theta$ as the parameters) $v_\theta^{(i)}: \mathbb{R}^d \to \mathbb{R}^d$ takes an interpolated sample $x_{t_i} \in \mathbb{R}^d$ as input and predicts the velocity vector. For notational convenience, we write $v_\theta(x_t, t_i) := v_\theta^{(i)}(x_t)$.

Each model $v_\theta^{(i)}$ is trained independently by minimizing the mean squared error between predicted and true velocity:

\begin{equation}
    \mathcal{L}^{(i)} = \mathbb{E}_{x \sim p_{\text{data}},\, z \sim \mathcal{N}(0, I_d)} \left[ \| v_\theta^{(i)}(x_{t_i}) - (x - z) \|_2^2 \right]
    \label{eq:Forest-Flow_loss}
\end{equation}
where $x_{t_i} = (1 - t_i) \cdot z + t_i \cdot x$ is the interpolated sample at time $t_i$. The expectation is approximated by pairing each data point with $K$ independent noise samples, creating $K \cdot n$ training pairs per time level, where $n$ is the number of samples in the training set.

\subsubsection{Scoring Functions for Forest-Flow}
All scores are computed in the preprocessed (scaled) space unless otherwise stated. Forest-Flow interpolates from noise ($t = 0$) to data ($t = 1$). We focus on later time levels for the Reconstruction and Deviation Score where the signal-to-noise ratio is higher (see formula of interpolation points):
\begin{equation}
    \mathcal{T}_{\text{Flow}} = \left\{ t_i : i \in \left\{ \lfloor n_t/2 \rfloor, \ldots, n_t - 2 \right\} \right\}
\end{equation}
The extreme index $i = n_t - 1$ (corresponding to $t = 1$) is excluded to avoid boundary effects where the interpolation reaches the data point exactly.

\paragraph{Decision Score.} The Decision (inspired by the original anomaly scoring function of TCCM) evaluates prediction accuracy at a single time level very close to the data distribution at $t_{n_t-2}$ (very high signal to noise ratio). For a test sample $x$ and $K$ independent noise samples $z^{(k)} \sim \mathcal{N}(0, I_d)$, the interpolated samples are:
\begin{equation}
    x_{t_{n_t-2}}^{(k)} = (1 - t_{n_t-2}) \cdot z^{(k)} + t_{n_t-2} \cdot x
\end{equation}
The Decision Score averages squared velocity prediction errors over all noise samples:
\begin{equation}
    s_{\text{Decision}}^{\text{Flow}}(x) = \frac{1}{K} \sum_{k=1}^{K} \left\| (x - z^{(k)}) - v_\theta\left(x_{t_{n_t-2}}^{(k)}, t_{n_t-2}\right) \right\|_2^2
    \label{eq:Forest-Flow_Decision}
\end{equation}
Normal samples should yield small errors because the model has learned accurate velocity predictions for in-distribution samples. Anomalous samples produce larger errors as they lie outside the learned flow trajectories.

\paragraph{Deviation Score.} The Deviation Score extends the Decision Score by accumulating prediction errors across multiple time levels in $\mathcal{T}_{\text{Flow}}$, providing a more comprehensive assessment of model fit along the trajectory:
\begin{equation}
    s_{\text{Deviation}}^{\text{Flow}}(x) = \frac{1}{K} \sum_{k=1}^{K} \sum_{i=\lfloor n_t/2 \rfloor}^{n_t-2} \left\| (x - z^{(k)}) - v_\theta\left(x_{t_i}^{(k)}, t_i\right) \right\|_2^2
    \label{eq:Forest-Flow_Deviation}
\end{equation}
where $x_{t_i}^{(k)} = (1 - t_i) \cdot z^{(k)} + t_i \cdot x$. The accumulation over multiple time levels amplifies systematic prediction errors for anomalous samples while averaging out random fluctuations for normal samples.

\paragraph{Reconstruction Score.} The Reconstruction Score measures how accurately the model reconstructs the original data by following the learned vector field over the full trajectory from intermediate points to the final time $t=1$. Starting from an interpolated point $x_{t_i}^{(k)}$, we solve the
Ordinary Differential Equation (ODE) forward using Euler integration
with step size $h = 1/(n_t - 1)$. We initialize
\begin{equation}
    x_0^{(i,k)} = x_{t_i}^{(k)}, \quad \tau_0 = t_i
\end{equation}
and iteratively update:
\begin{equation}
    x_{j+1}^{(i,k)} = x_j^{(i,k)} + h \cdot v_\theta\!\left(x_j^{(i,k)}, \tau_j\right),
    \quad
    \tau_{j+1} = \tau_j + h.
\end{equation}

Here, $j$ indexes the Euler integration steps, i.e., $j = 0,\ldots,(n_t-1)-i-1$, so that after $(n_t-1)-i$ steps the time reaches $\tau_{(n_t-1)-i} = 1$.
 The Reconstruction Score is computed in the original data space after applying inverse preprocessing transformations:
\begin{equation}
    s_{\text{Reconstruction}}^{\text{Flow}}(x) = \frac{1}{K} \sum_{k=1}^{K} \sum_{i=\lfloor n_t/2 \rfloor}^{n_t-2} \left\| \tilde{x} - \tilde{x}_{\text{end}}^{(i,k)} \right\|_2^2
    \label{eq:Forest-Flow_Reconstruction}
\end{equation}
where $\tilde{\cdot}$ denotes the post-processed sample in original data space and $x_{\text{end}}^{(i,k)} = x_{(n_t-1)-i}^{(i,k)}$. Small prediction errors can compound during integration, making it sensitive to systematic model errors but also more computationally demanding. This score measures the environment of the
learned vector field around the ideal trajectory of the test sample. Anomalous samples
should have a more disturbed vector field around the ideal trajectory.

\subsection{TCCM (Time-Conditioned Contraction Matching)}
\label{subsec:tccm}
TCCM \cite{li2025scalable} is a neural network-based flow matching method \cite{lipman2022flow} specifically designed for anomaly detection.

\subsubsection{Framework}
TCCM~\cite{TCCMCode, li2025scalable} learns a continuous time-conditioned contraction field that maps normal data toward the origin \cite{li2025scalable}.
Importantly, although this behavior can be interpreted as a contraction dynamic, the method is not based
on explicitly solving an ODE trajectory\cite{li2025scalable}. Instead, it uses a simplified supervision scheme with a fixed target
direction $-z$ for each training sample $z$ in all time steps.

Specifically, for $x \sim p_{\text{data}}$ and $t \sim \mathcal{U}(0,1)$, the model takes the augmented input
$\tilde x = [x; \mathrm{Embed}(t)]$ and predicts a velocity vector $f_\theta(\tilde x)$.  $f_\theta$ is the MLP with $\theta$ as its learned parameters. The time is encoded with sinusoidal embeddings \cite{vaswani2017attention}. Training minimizes
\begin{equation}
    \mathcal{L}_{\text{TCCM}}
    = \mathbb{E}_{x,t}\left[\left\| f_\theta([x;\mathrm{Embed}(t)]) + x \right\|_2\right].
    \label{eq:tccm_loss}
\end{equation}

\paragraph{Key Differences from Forest-Flow.} TCCM differs from Forest-Flow in three main aspects: (1) TCCM learns a contraction velocity toward the origin  while Forest-Flow learns a trajectory (ODE) between Gaussian and data distributions; TCCM is not a generative model (2) TCCM uses continuous time conditioning during training while Forest-Flow trains on discrete time levels; (3) TCCM uses one neural network (Multi-Layer Perceptron) as a Universal Function Approximator (UFA) for all time steps, while Forest-Flow uses XGBoost regressors \cite{jolicoeur2024forestflow} for each discrete time step. Due to deterministic training and inference, TCCM does not require Monte Carlo estimation ($K = 1$ implicitly).

\subsubsection{Scoring Functions for TCCM}
\label{sec:tccm_anomaly_scoring}
While TCCM is trained with continuous time \cite{li2025scalable}, our Deviation and Reconstruction Scores require discrete time levels for evaluation:
\begin{equation}
    t_i = \frac{i}{n_t - 1}, \quad i = 0, 1, \ldots, n_t - 1
    \label{eq:tccm_timelevels}
\end{equation}
TCCM time levels start at $t_0 = 0$. For the Deviation and Reconstruction Scores, we construct a synthetic linear interpolation path toward the origin:
\begin{equation}
    x(t_i) = (1 - t_i) \cdot x
    \label{eq:tccm_interpolation}
\end{equation}
This path is \emph{not} a true  ODE (Ordinary Differential Equation) trajectory (the model is not trained on that), but serves as a probe path for evaluation. For the Reconstruction and Deviation Score we use the first half of time levels where probe points remain closer to the original sample:
\begin{equation}
    \mathcal{T}_{\text{TCCM}} = \left\{ t_i : i \in \left\{ 0, \ldots, \lfloor n_t/2 \rfloor - 1 \right\} \right\}
\end{equation}
We do this to be consistent with the Forest-Flow approach. Additionally, this seems to enhance detection performance (sparsely tested) but is, unlike Forest-Flow, not connected to a higher signal-to-noise ratio (there is no noise included at all).

Instead, enhanced performance might come from the Lipschitz continuity of $f_{\theta}$ (the MLP, see below) \cite{li2025scalable}: Near the original data points, the Lipschitz constraint guaranties a well-behaved vector field, making the Deviation Score more sensitive to actual anomalies rather than numerical instabilities that could arise further along the interpolation path. In general, the Lipschitz continuity is very critical by ensuring robustness guaranties under input perturbation \cite{li2025scalable}. Note that \cite{li2025scalable} proves that the scoring function in the paper (Decision Score for TCCM) is Lipschitz continuous, whereas this property is not guaranteed for Deviation and Reconstruction Score.

Another property of TCCM is its explainability \cite{li2025scalable}, meaning that the anomalousness of a data sample can be traced back to a certain feature. This is possible because the residual vector itself encodes per-feature contributions to the anomaly score (Decision Score)\cite{li2025scalable}. Since residual vectors are calculated for Deviation and Decision Score as well (and then just averaged), these scoring function should also be interpretable.

\paragraph{Decision Score.} The Decision Score is the original anomaly detection function from~\cite{li2025scalable}. At $t = 1$, the ideal prediction for any point $x$ is $f_\theta(x, 1) = -x$. Any other time between $(0, 1]$ could have been chosen \cite{li2025scalable}. The Decision Score measures the deviation from this ideal:
\begin{equation}
    s_{\text{Decision}}^{\text{TCCM}}(x) = \| f_{\theta}\!\bigl([x;\mathrm{Embed}(1)\bigr]) + x \|_2
    \label{eq:tccm_Decision}
\end{equation}

Samples that don't fit into the data distribution (anomalies) should yield a higher deviation \cite{li2025scalable}.
Note that this uses the Euclidean norm, unlike all the other scores, which use squared errors.

\paragraph{Deviation Score.} The Deviation Score probes temporal consistency of the learned contraction field at multiple interpolated positions along a synthetic path toward the origin:
\begin{equation}
    s_{\text{Deviation}}^{\text{TCCM}}(x) = \sum_{i=0}^{\lfloor n_t/2 \rfloor - 1} \left\| f_{\theta}\!\bigl([x(t_i);\mathrm{Embed}(t_i)\bigr])
 + x(t_i) \right\|_2^2
    \label{eq:tccm_Deviation}
\end{equation}
where $x(t_i) = (1 - t_i) \cdot x$. At each interpolated point, the true contraction vector should be $-x(t_i)$. For normal samples, interpolated points tend to stay within regions where the contraction rule $f_\theta(z, t) \approx -z$ is well-supported, yielding consistently small residuals. For anomalous samples, the same probe points are more likely to fall into poorly covered regions, producing larger accumulated residuals.

\paragraph{Reconstruction Score.}
The Reconstruction Score measures how accurately the learned flow transports interpolated points toward the origin when integrated forward.
Starting from $x_0^{(i)} = x(t_i)$ at each time level $t_i \in \mathcal{T}_{\text{TCCM}}$ with $\tau_0=t_i$, we integrate using Euler's method:
\begin{equation}
x_{j+1}^{(i)} = x_j^{(i)} + h \cdot f_{\theta}\!\bigl(x_j^{(i)};\mathrm{Embed}(\tau_j)\bigr),
\quad
\tau_{j+1} = \tau_j + h,
\quad
h = \frac{1}{n_t - 1},
\label{eq:euler_tccm}
\end{equation}
with $j = 0,\ldots,(n_t-1)-i-1$.
After $(n_t-1)-i$ steps, the trajectory reaches $\tau=1$, and we denote the terminal state by
\begin{equation}
x_{\text{end}}^{(i)} := x_{(n_t-1)-i}^{(i)}.
\end{equation}
Because the intended terminal state at $\tau = 1$ is the origin, the Reconstruction Score accumulates endpoint deviations:
\begin{equation}
s_{\text{Reconstruction}}^{\text{TCCM}}(x) =
\sum_{i=0}^{\lfloor n_t/2 \rfloor - 1}
\left\| x_{\text{end}}^{(i)} \right\|_2^2.
\label{eq:tccm_Reconstruction}
\end{equation}

For normal samples, the learned flow transports the interpolated points closer to the origin. For anomalous samples, small local prediction inaccuracies compound, leading to systematic endpoint deviations.

\section{Related Work}

Unsupervised outlier detection (UOD) aims to identify anomalous samples without access to labeled data during training. Most deep UOD methods assume clean training sets~\citep{ruff2018deep, schlegl2017unsupervised}, an assumption that is frequently violated in large-scale scenarios where anomalies are inevitably present. This mismatch has motivated approaches that explicitly learn from contaminated datasets~\citep{Zhou2017RDA,zong2018deep}. As noted by \citep{huang2024entropystop}, performance degradation under training-set contamination is a well-known issue. Accordingly, prior work distinguishes two settings for unlabeled anomaly detection. The semi-supervised setting assumes access to uncontaminated data containing only normal samples (e.g., DeepSVDD~\cite{ruff2018deep}, NeuTraL AD~\cite{Qiu2021NeuTraLAD}, ICL~\cite{shenkar2022anomaly}, AnoGAN~\cite{schlegl2017unsupervised}). In contrast, the unsupervised setting operates directly on contaminated datasets, identifying anomalies within the training data or via adaptation to a test set (e.g., RandNet~\cite{Chen2017RandNet}, ROBOD~\cite{Ding2022ROBOD}, RDA~\cite{Zhou2017RDA}). State-of-the-art deep unsupervised methods typically rely on ensemble strategies to improve robustness against contamination~\cite{Chen2017RandNet, Ding2022ROBOD,Liu2019GenerativeAdversarial,Wang2019RandomDistances}. These approaches are not considered in this work, as our focus lies on flow-matching models. The recently proposed approach TCCM~\citep{li2025scalable}, which uses a specialized flow-matching for semi-supervised anomaly detection, includes a contamination ablation, observing uniform performance degradation as contamination increases. However, it does not investigate whether alternative anomaly scoring strategies or conventional flow-matching formulations could mitigate this effect.

Flow-matching has emerged as a promising generative modeling framework, combining the training stability and expressivity of diffusion models with significantly lower computational cost~\cite{liu2022flow_straight_fast,lee2023minimizing_trajectory_curvature,li2025scalable}. It learns an ordinary differential equation (ODE) that deterministically maps samples from a source to a target distribution, enabling faster sampling and simpler optimization compared to diffusion models based on stochastic differential equations (SDEs)~\cite{li2025scalable,livernoche2023diffusion}. Unlike diffusion models, flow-matching does not require a forward noising process, nor does it rely on explicit density estimation as in normalizing flows~\cite{kobyzev2020normalizing}. New samples are generated by evolving draws from the source distribution along the learned ODE toward the target distribution.

For tabular data, Forest-Flow~\citep{jolicoeur2024forestflow} is a generative flow-matching model that predicts velocity fields at discrete time steps using XGBoost regressors and achieves strong performance on generation benchmarks. TCCM~\citep{li2025scalable}, a recent anomaly detection method, adapts flow matching for anomaly detection and differs from Forest-Flow in three key aspects: (i) TCCM learns a contraction velocity that maps data samples toward the origin rather than a generative trajectory from a Gaussian source to the data distribution; (ii) TCCM employs continuous-time conditioning during training, whereas Forest-Flow operates on discrete time levels; and (iii) TCCM uses a single shared multi-layer perceptron across all time steps, while Forest-Flow trains separate XGBoost regressors per time step. Due to its deterministic formulation, TCCM does not require Monte Carlo estimation ($K=1$). While TCCM demonstrates strong empirical performance, alternative scoring functions beyond the original formulation have not yet been explored within its architecture.

\section{Synthetic Business Transaction Dataset}
\label{app:synthetic_dataset}

\begin{table}[ht]
\centering
\small
\caption{Feature groups in the Synthetic Business Transaction dataset.}
\label{tab:feature_groups}
\begin{tabular}{@{}lp{0.62\linewidth}@{}}
\toprule
\textbf{Group} & \textbf{Features} \\
\midrule
Amount & \texttt{amount}, \texttt{amount\_zscore\_series}, \texttt{amount\_ratio\_to\_mean} \\
\addlinespace
Timing & \texttt{day\_of\_month}, \texttt{day\_of\_week}, \texttt{month}, \texttt{days\_since\_last\_in\_series}, \texttt{day\_deviation\_from\_usual} \\
\addlinespace
Mismatch flags & \texttt{iban\_mismatch}, \texttt{swift\_mismatch}, \texttt{method\_mismatch}, \texttt{channel\_mismatch}, \texttt{known\_name\_new\_iban}, \texttt{is\_card\_mobile} \\
\addlinespace
Novelty & \texttt{is\_new\_series}, \texttt{is\_new\_ref\_name}, \texttt{is\_new\_iban} \\
\addlinespace
History & \texttt{series\_tx\_count\_before\_log}, \texttt{ref\_name\_count\_before\_log}, \texttt{iban\_count\_before\_log}, \texttt{tx\_count\_this\_month\_so\_far} \\
\bottomrule
\end{tabular}
\end{table}

This section specifies the generation procedure, preprocessing, and evaluation split of the Synthetic Business Transaction dataset.

\paragraph{Overview.}
The dataset models business-to-business (B2B) payment activity for nine customers over a fixed time period. It comprises $6{,}364$ transactions with $50$ labeled anomalies, corresponding to an anomaly ratio $\rho = 50/6364 \approx 0.786\%$, and is constructed to assess detection performance under extreme class imbalance.

\paragraph{Nominal behavior.}
Each customer is associated with one primary recurring payment series to a fixed business partner, together with five to ten additional background series to further partners. Each series consists of monthly payments whose amounts follow a Gaussian distribution around a series-specific base value, with per-series payment terms and mild multiplicative growth or decline over time. In addition, each customer issues sporadic one-off payments. These regularities define the nominal distribution against which anomalies are measured.

\paragraph{Anomaly injection.}
Anomalies are drawn from seven types: amount spikes ($1.5$--$2\times$ the expected value), duplicate payments within a series, modified payee information, temporal shifts, atypical payment channels (card/mobile in place of the usual method), IBAN inconsistencies relative to series history, and subtle payee-name perturbations accompanied by an IBAN change (e.g., ``Solutions'' $\rightarrow$ ``Systems''). For each customer, a single anomaly type is selected at random and applied a random number of times to that customer's primary series; the remaining series are left nominal. This per-customer procedure yields a total of $50$ anomalies across the dataset.

\paragraph{Preprocessing.}
Each transaction is encoded by $21$ numerical features ~\ref{tab:feature_groups}. All features are computed in temporal order, such that the representation of a transaction depends solely on preceding transactions; mismatch flags, in particular, compare the current value against the modal historical value of the same series. This sequential construction precludes information leakage from future observations.

\paragraph{Training and Evaluation split.}
The data is partitioned chronologically at its temporal midpoint (by day), yielding $3{,}188$ training transactions ($28$ anomalies) and $3{,}176$ test transactions ($22$ anomalies). 

\definecolor{ffxgbcolor}{RGB}{214,39,40}
\definecolor{tccmcolor}{RGB}{44,160,44}

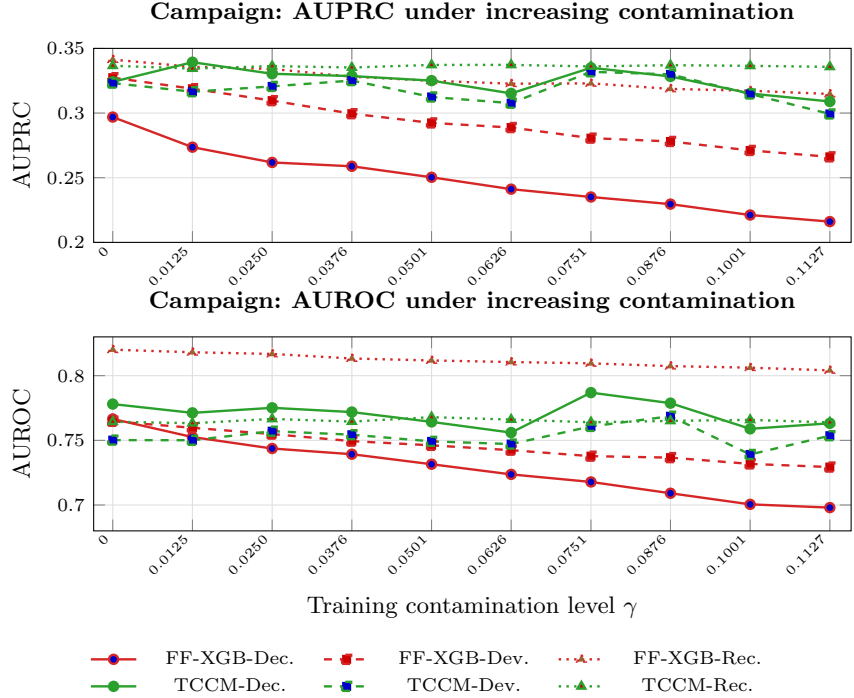
\begin{figure}[H]
\centering

\pgfplotstableread{
cont FFXGBDecAUPRC FFXGBDevAUPRC FFXGBRecAUPRC TCCMDecAUPRC TCCMDevAUPRC TCCMRecAUPRC FFXGBDecAUROC FFXGBDevAUROC FFXGBRecAUROC TCCMDecAUROC TCCMDevAUROC TCCMRecAUROC
0 0.29691 0.327346 0.341343 0.324149 0.323156 0.336589 0.766545 0.764736 0.820098 0.778046 0.750187 0.764403
0.012517301242001 0.273634 0.318827 0.335739 0.339324 0.316504 0.334383 0.752539 0.759700 0.818098 0.771248 0.750081 0.763240
0.025042602484003 0.261827 0.309617 0.334108 0.330380 0.320632 0.336330 0.743688 0.754826 0.816807 0.775147 0.757077 0.766553
0.037561903726004 0.258845 0.299523 0.328158 0.328592 0.325145 0.335227 0.739278 0.749393 0.813280 0.771869 0.754361 0.764503
0.050085204968006 0.250352 0.292392 0.324966 0.325091 0.312453 0.337223 0.731528 0.746140 0.811764 0.764271 0.749214 0.767915
0.062586506210007 0.241164 0.288802 0.322720 0.315193 0.307660 0.337245 0.723684 0.742388 0.810546 0.755990 0.747114 0.766047
0.075127807452009 0.235105 0.280628 0.322972 0.335004 0.331885 0.336055 0.717841 0.737776 0.809526 0.786903 0.760779 0.763962
0.08761910869401 0.229594 0.278058 0.318680 0.328321 0.329940 0.337016 0.709097 0.736648 0.807397 0.778754 0.768639 0.765040
0.10013740993601 0.221165 0.271103 0.317277 0.315139 0.314772 0.336517 0.700520 0.731784 0.806192 0.758915 0.739085 0.765814
0.1126547117801 0.216080 0.266162 0.314759 0.308961 0.299321 0.335727 0.697990 0.729350 0.804098 0.763178 0.753605 0.763887
}\campaigncontamdata

\begin{tikzpicture}
\begin{groupplot}[
    group style={group size=1 by 2, vertical sep=1.25cm},
    width=0.95\textwidth,
    height=0.34\textwidth,
    xmin=-0.003,
    xmax=0.116,
    xtick={0,0.012517301242001,0.025042602484003,0.037561903726004,0.050085204968006,0.062586506210007,0.075127807452009,0.08761910869401,0.10013740993601,0.1126547117801},
    xticklabels={0,0.0125,0.0250,0.0376,0.0501,0.0626,0.0751,0.0876,0.1001,0.1127},
    x tick label style={font=\tiny, rotate=45, anchor=east},
    y tick label style={font=\scriptsize},
    label style={font=\small},
    title style={font=\small\bfseries},
    ymajorgrids=true,
    xmajorgrids=true,
    grid style={draw=gray!25},
    every axis plot/.append style={line width=0.9pt, mark size=1.7pt},
]

\nextgroupplot[
    title={Campaign: AUPRC under increasing contamination},
    ylabel={AUPRC},
    ymin=0.20,
    ymax=0.35,
    legend to name=campaignlegend,
    legend columns=3,
    legend style={font=\scriptsize, draw=none, column sep=0.3cm}
]

\addplot+[color=ffxgbcolor, mark=*] table[x=cont,y=FFXGBDecAUPRC]{\campaigncontamdata};
\addlegendentry{FF-XGB-Dec.}

\addplot+[color=ffxgbcolor, dashed, mark=square*] table[x=cont,y=FFXGBDevAUPRC]{\campaigncontamdata};
\addlegendentry{FF-XGB-Dev.}

\addplot+[color=ffxgbcolor, dotted, mark=triangle*] table[x=cont,y=FFXGBRecAUPRC]{\campaigncontamdata};
\addlegendentry{FF-XGB-Rec.}

\addplot+[color=tccmcolor, mark=*] table[x=cont,y=TCCMDecAUPRC]{\campaigncontamdata};
\addlegendentry{TCCM-Dec.}

\addplot+[color=tccmcolor, dashed, mark=square*] table[x=cont,y=TCCMDevAUPRC]{\campaigncontamdata};
\addlegendentry{TCCM-Dev.}

\addplot+[color=tccmcolor, dotted, mark=triangle*] table[x=cont,y=TCCMRecAUPRC]{\campaigncontamdata};
\addlegendentry{TCCM-Rec.}

\nextgroupplot[
    title={Campaign: AUROC under increasing contamination},
    xlabel={Training contamination level $\gamma$},
    ylabel={AUROC},
    ymin=0.68,
    ymax=0.83
]

\addplot+[color=ffxgbcolor, mark=*] table[x=cont,y=FFXGBDecAUROC]{\campaigncontamdata};
\addplot+[color=ffxgbcolor, dashed, mark=square*] table[x=cont,y=FFXGBDevAUROC]{\campaigncontamdata};
\addplot+[color=ffxgbcolor, dotted, mark=triangle*] table[x=cont,y=FFXGBRecAUROC]{\campaigncontamdata};

\addplot+[color=tccmcolor, mark=*] table[x=cont,y=TCCMDecAUROC]{\campaigncontamdata};
\addplot+[color=tccmcolor, dashed, mark=square*] table[x=cont,y=TCCMDevAUROC]{\campaigncontamdata};
\addplot+[color=tccmcolor, dotted, mark=triangle*] table[x=cont,y=TCCMRecAUROC]{\campaigncontamdata};

\end{groupplot}
\end{tikzpicture}

\vspace{0.5em}
\pgfplotslegendfromname{campaignlegend}

\caption{Campaign contamination curves for AUPRC and AUROC. Campaign contains
41{,}188 samples with 4{,}640 anomalies, giving an anomaly ratio
$\rho = 0.1127$ or $11.27\%$. The x-axis reports the training contamination
level $\gamma$ as a fraction, where $\gamma=0$ means anomaly-free training data,
$\gamma=0.0125$ means approximately $1.25\%$ contamination, and
$\gamma=0.1127 \approx \rho$ corresponds to full contamination. We compare TCCM
with Forest-Flow using XGBoost, and plot Decision, Deviation, and Reconstruction
anomaly scores using the mean values from the Campaign table.}
\label{fig:campaign_contamination_curves}
\end{figure}

\section{Additional Contamination Curves}~\label{contamination}
Figures~\ref{fig:campaign_contamination_curves},
\ref{fig:business_contamination_curves}, and
\ref{fig:waveform_contamination_curves} show AUROC and AUPRC trends across all
training contamination levels for Campaign, Synthetic Business Transaction, and
Waveform, respectively. The x-axis denotes the contamination level $\gamma$,
where $\gamma=0$ indicates clean training data and the final point corresponds to
the full-contamination setting. For each dataset, we compare TCCM, Forest-Flow
with XGBoost, using Decision, Deviation, and
Reconstruction scores, complementing Table~\ref{tab:four_dataset_compact_results} in main paper
with performance trends beyond the zero- and full-contamination endpoints.



\definecolor{ffxgbcolor}{RGB}{214,39,40}
\definecolor{tccmcolor}{RGB}{44,160,44}

\begin{figure}[H]
\centering

\pgfplotstableread{
cont FFXGBDecAUPRC FFXGBDevAUPRC FFXGBRecAUPRC TCCMDecAUPRC TCCMDevAUPRC TCCMRecAUPRC FFXGBDecAUROC FFXGBDevAUROC FFXGBRecAUROC TCCMDecAUROC TCCMDevAUROC TCCMRecAUROC
0 0.036535 0.19744 0.059194 0.036986 0.032955 0.073751 0.661573 0.777540 0.824624 0.697040 0.678668 0.869225
0.0009759 0.009366 0.019823 0.117421 0.009831 0.008365 0.067905 0.594771 0.749703 0.832340 0.596677 0.561688 0.848135
0.0019518 0.008716 0.018828 0.112020 0.009466 0.008387 0.067082 0.565236 0.733343 0.838168 0.587692 0.559287 0.846582
0.0029276 0.008090 0.016347 0.111158 0.009480 0.008335 0.066751 0.547513 0.720150 0.830293 0.578097 0.518070 0.846045
0.0039035 0.008682 0.016818 0.108611 0.008135 0.007516 0.064966 0.566876 0.723079 0.821903 0.540630 0.521600 0.844374
0.0048794 0.008114 0.016199 0.107613 0.008161 0.007445 0.065883 0.553516 0.722834 0.822771 0.538678 0.515308 0.846558
0.0058553 0.008324 0.015840 0.104105 0.008072 0.007491 0.065341 0.551727 0.716107 0.812734 0.541157 0.519496 0.846365
0.0068312 0.008295 0.015105 0.100966 0.007744 0.007268 0.065592 0.543440 0.705892 0.804038 0.526797 0.504641 0.846380
0.0078071 0.008423 0.014597 0.098961 0.008290 0.007843 0.064364 0.546584 0.701092 0.806774 0.545039 0.534626 0.846063
0.0087829 0.007839 0.013322 0.093797 0.007005 0.006821 0.064394 0.523338 0.686421 0.788770 0.488419 0.476350 0.845541
}\businesscontamdata

\begin{tikzpicture}
\begin{groupplot}[
    group style={group size=1 by 2, vertical sep=1.25cm},
    width=0.95\textwidth,
    height=0.34\textwidth,
    xmin=-0.00025,
    xmax=0.00905,
    xtick={0,0.0009759,0.0019518,0.0029276,0.0039035,0.0048794,0.0058553,0.0068312,0.0078071,0.0087829},
    xticklabels={0,0.0010,0.0020,0.0029,0.0039,0.0049,0.0059,0.0068,0.0078,0.0088},
    x tick label style={font=\tiny, rotate=45, anchor=east},
    y tick label style={font=\scriptsize},
    label style={font=\small},
    title style={font=\small\bfseries},
    ymajorgrids=true,
    xmajorgrids=true,
    grid style={draw=gray!25},
    every axis plot/.append style={line width=0.9pt, mark size=1.7pt},
]

\nextgroupplot[
    title={Business: AUPRC under increasing contamination},
    ylabel={AUPRC},
    ymin=0,
    ymax=0.21,
    legend to name=businesslegend,
    legend columns=3,
    legend style={font=\scriptsize, draw=none, column sep=0.3cm}
]

\addplot+[color=ffxgbcolor, mark=*] table[x=cont,y=FFXGBDecAUPRC]{\businesscontamdata};
\addlegendentry{FF-XGB-Dec.}

\addplot+[color=ffxgbcolor, dashed, mark=square*] table[x=cont,y=FFXGBDevAUPRC]{\businesscontamdata};
\addlegendentry{FF-XGB-Dev.}

\addplot+[color=ffxgbcolor, dotted, mark=triangle*] table[x=cont,y=FFXGBRecAUPRC]{\businesscontamdata};
\addlegendentry{FF-XGB-Rec.}

\addplot+[color=tccmcolor, mark=*] table[x=cont,y=TCCMDecAUPRC]{\businesscontamdata};
\addlegendentry{TCCM-Dec.}

\addplot+[color=tccmcolor, dashed, mark=square*] table[x=cont,y=TCCMDevAUPRC]{\businesscontamdata};
\addlegendentry{TCCM-Dev.}

\addplot+[color=tccmcolor, dotted, mark=triangle*] table[x=cont,y=TCCMRecAUPRC]{\businesscontamdata};
\addlegendentry{TCCM-Rec.}

\nextgroupplot[
    title={Business: AUROC under increasing contamination},
    xlabel={Training contamination level $\gamma$},
    ylabel={AUROC},
    ymin=0.45,
    ymax=0.89
]

\addplot+[color=ffxgbcolor, mark=*] table[x=cont,y=FFXGBDecAUROC]{\businesscontamdata};
\addplot+[color=ffxgbcolor, dashed, mark=square*] table[x=cont,y=FFXGBDevAUROC]{\businesscontamdata};
\addplot+[color=ffxgbcolor, dotted, mark=triangle*] table[x=cont,y=FFXGBRecAUROC]{\businesscontamdata};

\addplot+[color=tccmcolor, mark=*] table[x=cont,y=TCCMDecAUROC]{\businesscontamdata};
\addplot+[color=tccmcolor, dashed, mark=square*] table[x=cont,y=TCCMDevAUROC]{\businesscontamdata};
\addplot+[color=tccmcolor, dotted, mark=triangle*] table[x=cont,y=TCCMRecAUROC]{\businesscontamdata};

\end{groupplot}
\end{tikzpicture}

\vspace{0.5em}
\pgfplotslegendfromname{businesslegend}

\caption{Business dataset contamination curves for AUPRC and AUROC. The
Business dataset contains 6{,}364 samples with 50 anomalies, corresponding to a
dataset-level anomaly ratio of $\rho = 50/6364 \approx 0.00786$ or $0.786\%$.
The x-axis reports the training contamination level $\gamma$ as a fraction,
where $\gamma=0$ means anomaly-free training data and, for example,
$\gamma=0.0009759$ means approximately $0.0976\%$ contamination. The final
plotted level, $\gamma=0.0087829$, corresponds to the full-contamination setting
used in this training split. We compare TCCM with Forest-Flow using XGBoost, and
plot Decision, Deviation, and Reconstruction anomaly scores using the mean values
from the Business table.}
\label{fig:business_contamination_curves}
\end{figure}
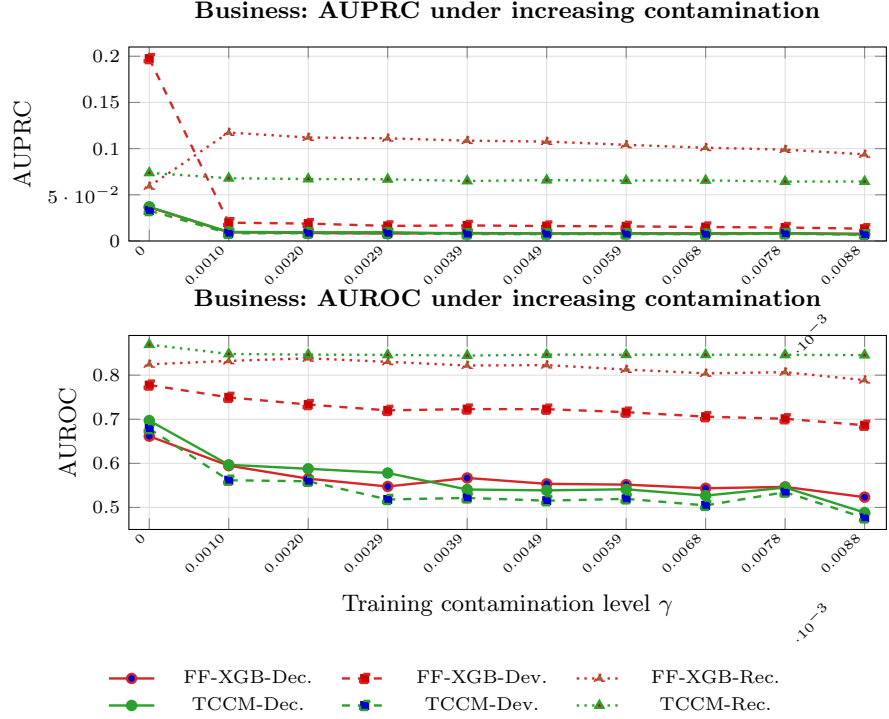



\definecolor{ffxgbcolor}{RGB}{214,39,40}
\definecolor{tccmcolor}{RGB}{44,160,44}

\begin{figure}[H]
\centering

\pgfplotstableread{
cont FFXGBDecAUPRC FFXGBDevAUPRC FFXGBRecAUPRC TCCMDecAUPRC TCCMDevAUPRC TCCMRecAUPRC FFXGBDecAUROC FFXGBDevAUROC FFXGBRecAUROC TCCMDecAUROC TCCMDevAUROC TCCMRecAUROC
0 0.068788 0.080000 0.156264 0.055606 0.067812 0.043311 0.690911 0.734957 0.741182 0.676816 0.676349 0.612938
0.0032281 0.066545 0.078479 0.152875 0.051743 0.061376 0.043401 0.678859 0.728045 0.735821 0.663967 0.679222 0.620653
0.0064562 0.061635 0.076151 0.140423 0.046707 0.050011 0.043660 0.667392 0.713768 0.725775 0.624337 0.603033 0.622581
0.0096843 0.053119 0.069503 0.137679 0.048123 0.060576 0.042601 0.640208 0.694148 0.725316 0.665476 0.699775 0.612684
0.0129124 0.049329 0.063386 0.131652 0.044670 0.051670 0.042714 0.635483 0.680440 0.716947 0.645624 0.645955 0.610199
0.0161405 0.052403 0.065556 0.115763 0.046498 0.056088 0.043013 0.648335 0.701775 0.711813 0.653574 0.678316 0.615656
0.0193686 0.049877 0.063182 0.119400 0.041718 0.051367 0.041835 0.630519 0.698395 0.712467 0.592493 0.617057 0.603993
0.0225967 0.051376 0.059648 0.113189 0.035428 0.036128 0.042493 0.644337 0.681330 0.710242 0.530556 0.521318 0.611179
0.0258248 0.048216 0.062664 0.107140 0.042198 0.046466 0.042478 0.619612 0.681722 0.707024 0.601861 0.615703 0.611333
0.0290529 0.046035 0.059930 0.101832 0.042489 0.051672 0.043058 0.613538 0.677720 0.705864 0.596256 0.610471 0.616242
}\waveformcontamdata

\begin{tikzpicture}
\begin{groupplot}[
group style={group size=1 by 2, vertical sep=1.45cm},
width=0.95\linewidth,
height=0.36\linewidth,
xmin=-0.001,
xmax=0.030,
xtick={0,0.0032281,0.0064562,0.0096843,0.0129124,0.0161405,0.0193686,0.0225967,0.0258248,0.0290529},
xticklabels={0,0.0032,0.0065,0.0097,0.0129,0.0161,0.0194,0.0226,0.0258,0.0291},
x tick label style={font=\tiny, rotate=45, anchor=east},
y tick label style={font=\scriptsize},
label style={font=\small},
title style={font=\small\bfseries},
ymajorgrids=true,
xmajorgrids=true,
grid style={draw=gray!25},
every axis plot/.append style={line width=0.9pt, mark size=1.7pt},
]

\nextgroupplot[
title={Waveform: AUPRC under increasing contamination},
ylabel={AUPRC},
ymin=0.03,
ymax=0.17,
legend to name=waveformlegend,
legend columns=3,
legend style={font=\scriptsize, draw=none, column sep=0.3cm}
]

\addplot+[color=ffxgbcolor, mark=*] table[x=cont,y=FFXGBDecAUPRC]{\waveformcontamdata};
\addlegendentry{FF-XGB-Dec.}

\addplot+[color=ffxgbcolor, dashed, mark=square*] table[x=cont,y=FFXGBDevAUPRC]{\waveformcontamdata};
\addlegendentry{FF-XGB-Dev.}

\addplot+[color=ffxgbcolor, dotted, mark=triangle*] table[x=cont,y=FFXGBRecAUPRC]{\waveformcontamdata};
\addlegendentry{FF-XGB-Rec.}

\addplot+[color=tccmcolor, mark=*] table[x=cont,y=TCCMDecAUPRC]{\waveformcontamdata};
\addlegendentry{TCCM-Dec.}

\addplot+[color=tccmcolor, dashed, mark=square*] table[x=cont,y=TCCMDevAUPRC]{\waveformcontamdata};
\addlegendentry{TCCM-Dev.}

\addplot+[color=tccmcolor, dotted, mark=triangle*] table[x=cont,y=TCCMRecAUPRC]{\waveformcontamdata};
\addlegendentry{TCCM-Rec.}

\nextgroupplot[
title={Waveform: AUROC under increasing contamination},
xlabel={Training contamination level $\gamma$},
ylabel={AUROC},
ymin=0.50,
ymax=0.76
]

\addplot+[color=ffxgbcolor, mark=*] table[x=cont,y=FFXGBDecAUROC]{\waveformcontamdata};
\addplot+[color=ffxgbcolor, dashed, mark=square*] table[x=cont,y=FFXGBDevAUROC]{\waveformcontamdata};
\addplot+[color=ffxgbcolor, dotted, mark=triangle*] table[x=cont,y=FFXGBRecAUROC]{\waveformcontamdata};

\addplot+[color=tccmcolor, mark=*] table[x=cont,y=TCCMDecAUROC]{\waveformcontamdata};
\addplot+[color=tccmcolor, dashed, mark=square*] table[x=cont,y=TCCMDevAUROC]{\waveformcontamdata};
\addplot+[color=tccmcolor, dotted, mark=triangle*] table[x=cont,y=TCCMRecAUROC]{\waveformcontamdata};

\end{groupplot}
\end{tikzpicture}

\vspace{0.5em}
\pgfplotslegendfromname{waveformlegend}

\caption{Waveform contamination curves for AUPRC and AUROC. Waveform has an
anomaly ratio of approximately $\rho = 0.0291$ or $2.90\%$. The x-axis reports
the training contamination level $\gamma$ as a fraction, where $\gamma=0$ means
anomaly-free training data, $\gamma=0.0032$ means approximately $0.32\%$
contamination, and $\gamma=0.0291 \approx \rho$ corresponds to full
contamination. We compare TCCM with Forest-Flow using XGBoost, and plot
Decision, Deviation, and Reconstruction anomaly scores.}
\label{fig:waveform_contamination_curves}
\end{figure}
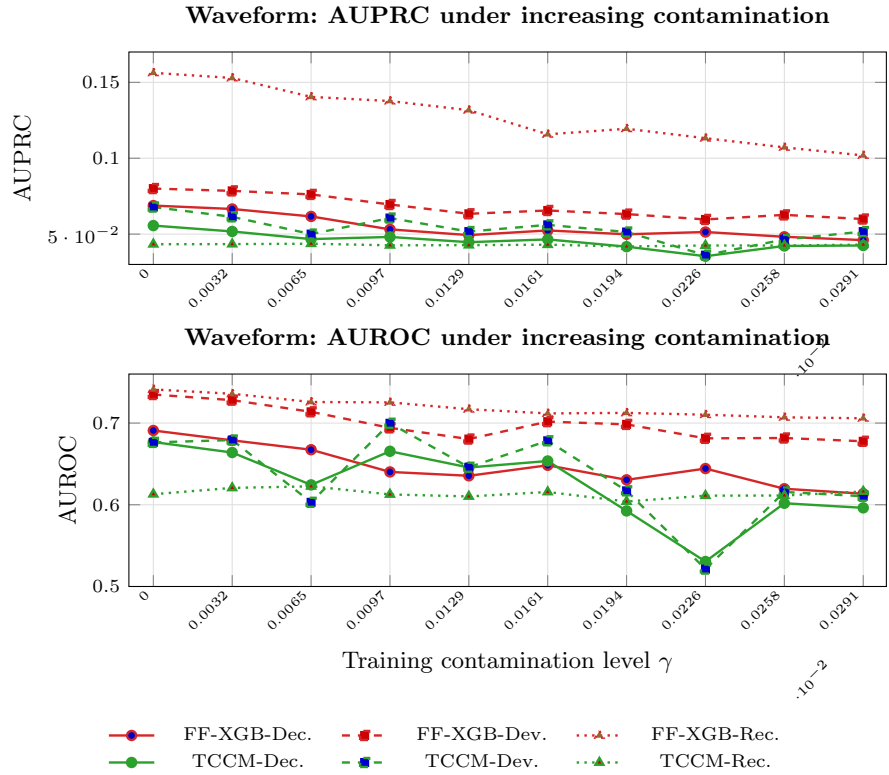

\clearpage
\bibliography{references}
\bibliographystyle{splncs04}

\end{document}